\pdfoutput=1

\documentclass[11pt]{article}

\usepackage[final]{acl}

\usepackage{times}
\usepackage{latexsym}

\usepackage[T1]{fontenc}

\usepackage[utf8]{inputenc}

\usepackage{microtype}

\usepackage{graphicx}
\usepackage{multicol}
\usepackage{booktabs}
\usepackage{amsmath}
\usepackage{upgreek}
\usepackage{footmisc}
\usepackage{tablefootnote}

\newlength{\bibitemsep}
\newlength{\bibparskip}
\let\oldthebibliography\thebibliography
\renewcommand\thebibliography[1]{%
  \oldthebibliography{#1}%
  \setlength{\parskip}{\bibitemsep}%
  \setlength{\itemsep}{\bibparskip}%
}

\title{VScript: Controllable Script Generation with Visual Presentation}

\author{Ziwei Ji, Yan Xu, I-Tsun Cheng, Samuel Cahyawijaya, Rita Frieske, \\   \large{\textbf{Etsuko Ishii, Min Zeng, Andrea Madotto, Pascale Fung}}\\
    Center for Artificial Intelligence Research (CAiRE)\\
    Hong Kong University of Science and Technology\\
  {zjiad@connect.ust.hk, pascale@ece.ust.hk}
	    }

\begin{document}

\maketitle
\begin{abstract}
In order to offer a customized script tool and inspire professional scriptwriters, we present \textbf{VScript}. It is a controllable pipeline that generates complete scripts, including dialogues and scene descriptions, as well as presents visually using video retrieval.  
With an interactive interface, our system allows users to select genres and input starting words that control the theme and development of the generated script. 
We adopt a hierarchical structure, which first generates the plot, then the script and its visual presentation.
A novel approach is also introduced to plot-guided dialogue generation by treating it as an inverse dialogue summarization. 
The experiment results show that our approach outperforms the baselines on both automatic and human evaluations, especially in genre control. 

\end{abstract}

\section{Introduction}
Artificial intelligence (AI) introduces significant changes in the creation of artworks, such as stylistic painting ~\citep{kotovenko2019content}, poem writing~\citep{hu2020generating}, music composition~\citep{dong2018musegan}, and converting the perception of creativity.
In particular, AI can assist in streamlining the art-making process and give humans fresh inspirations~\citep{anantrasirichai2021artificial}, and one plausible application is scriptwriting.
As a specific literary form, the script is indispensable in cinematography and theater~\citep{owens2012video,walker2012annotated}.
To enable the collaboration between scriptwriter and AI, an automatic script generation system must equip with three aspects.
First, the system must generate a complete script consisting of chronological scene descriptions and dialogues. 
Second, the system should provide controllability, e.g., customize script genre or storyline~\citep{cavazza2017introduction}.
Third, as a creative work, the generated script is required to have rich and diverse content.

\begin{figure}[!t]
 \centering
 \includegraphics[width=1\linewidth]{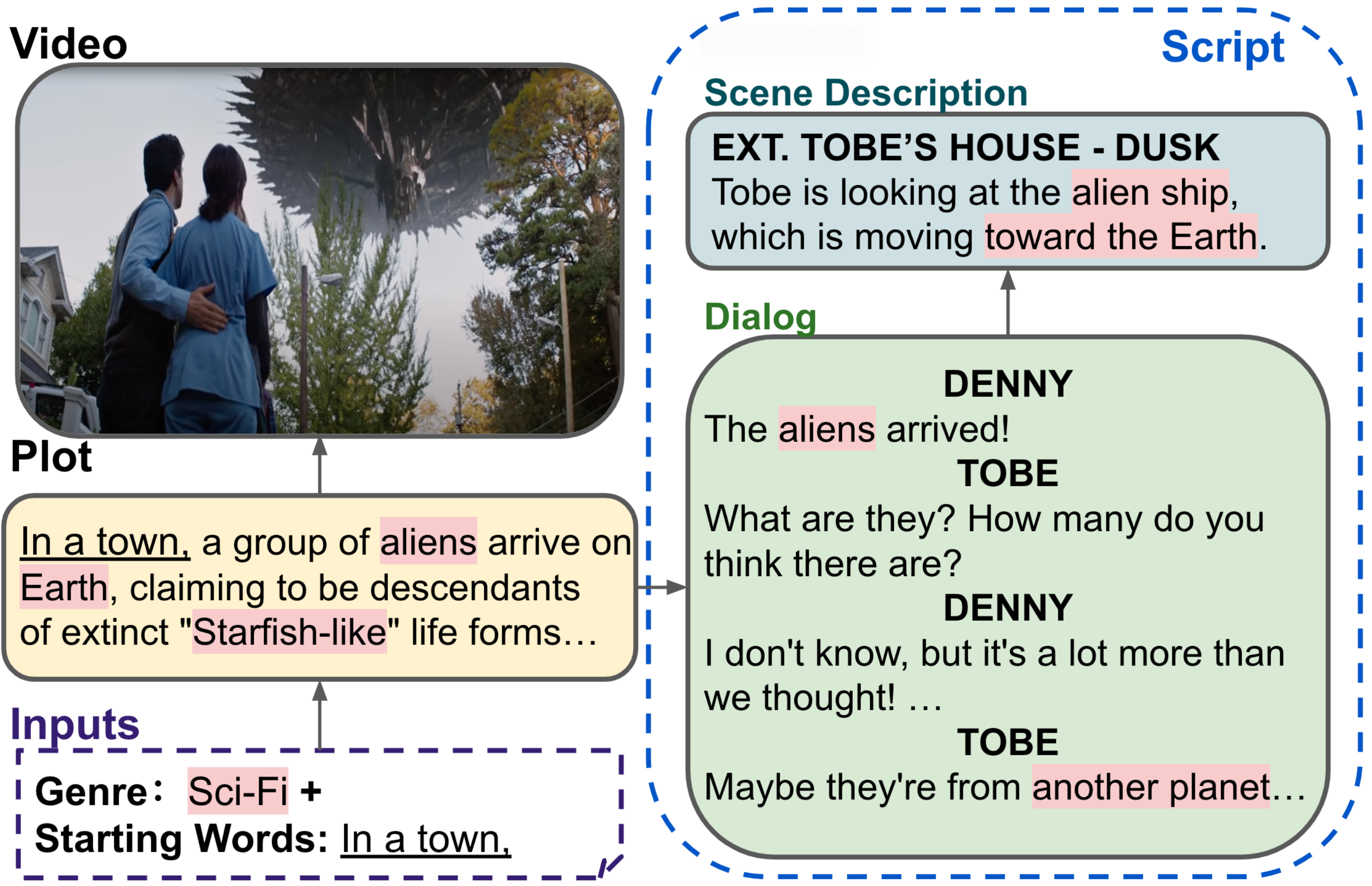}
  \caption{An example of the generated script (right) with its visual presentation (top left) from \textbf{VScript}.  
  Given the inputs, i.e., genre and starting words, a plot is generated, which guides the generation of a script consisting of a scene description and a dialogue. The words highlighted in pink show the belongingness to the given genre (\texttt{Sci-Fi}).
  Additionally, the video vividly presents the script.
  }
  \label{fig:example}
    \vspace{-1em}
\end{figure}

\begin{figure*}[!ht]
 \centering
 \includegraphics[width=0.9\linewidth]{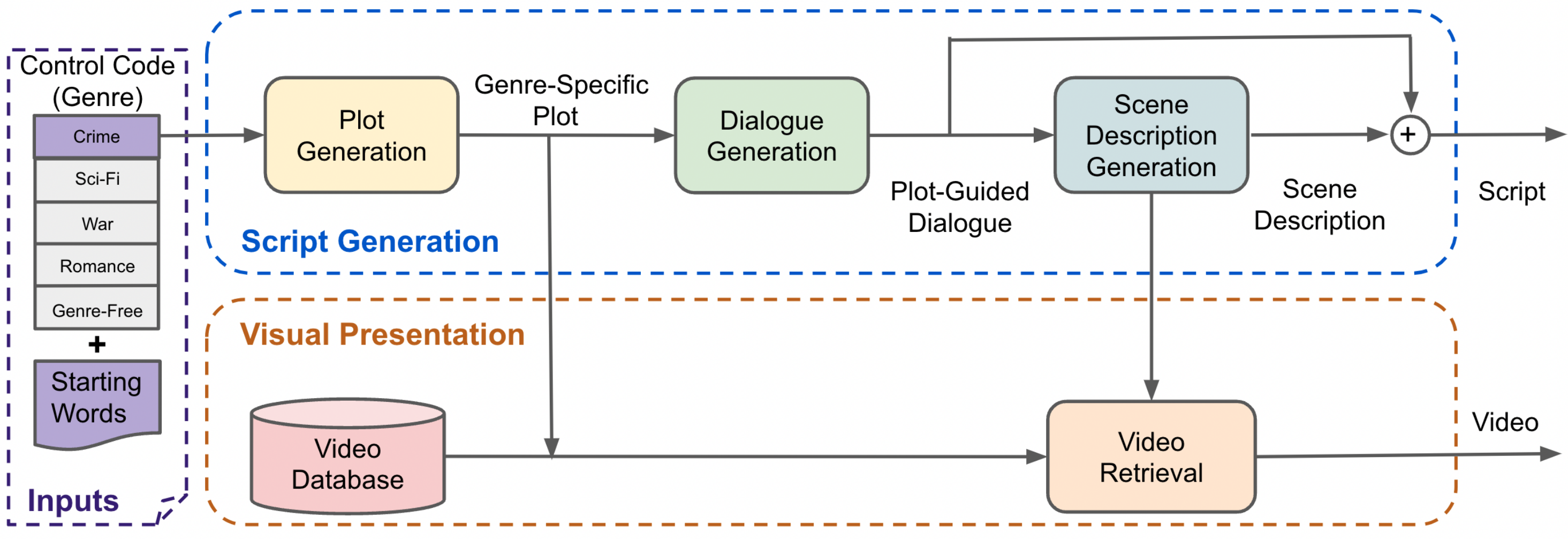}

  \caption{
  \textbf{VScript} consists of two modules, i.e., \textbf{Script Generation} and \textbf{Visual Presentation}. Given the genre and starting words, the Script Generation module generates a genre-specific plot, dialogues, and scene descriptions. The Visual Presentation module searches for a relevant visualization from a large video database. 
}
  \label{fig:system}
    \vspace{-1em}
\end{figure*}

While ~\citet{zhu2020scriptwriter} propose a narrative-guided script generation task, they only focus on retrieving dialogue utterances and omit scene description, an essential component of a script describing the environment and characters. 
Other works~\citep{chen2019neural,bensaid2021fairytailor} take inspiration from storyboarding and utilize a series of images to demonstrate stories. Nevertheless, no prior work presents scripts by leveraging video, a more informative and attractive medium.
In addition, prior works lack the ability to control certain elements, such as genre. Controllable generation systems~\citep{keskar2019ctrl, dathathri2019plug,madotto2020ppcm} can help to allow the customization of scripts based on user preferences.


In this paper, we present \textbf{VScript}, a controllable script generation system that includes all essential components of a script. We adopt a hierarchical structure for our framework by implementing high-level planning~\citep{fan-etal-2018-hierarchical} and following the guidelines of video-making processes~\citep{owens2012video}. 
As shown in Figure \ref{fig:system}, \textbf{VScript} is composed by \textbf{Script Generation} and \textbf{Visual Presentation} modules. 
The examples of corresponding components are in Figure \ref{fig:example}.
\textbf{Script Generation} module consists of three sub-modules: 1) plot generation, 2) dialogue generation, and 3) scene description generation.
Firstly, the system generates a genre-specific plot as an outline of the script. To avoid generating aimless scripts, the system allows users to provide some brief initial information, i.e., genre and the beginning of the plot, for a high degree of control over the script on the genre and story trend. 
Secondly, we conduct a zero-shot plot-guided dialogue generation by framing the problem as an inverse task of abstractive dialogue summarization.
Finally, we generate scene descriptions based on the dialogues to produce complete scripts. 
\textbf{Visual Presentation} module vividly demonstrates the script by retrieving videos from an automatically constructed video database. It can serve as a rough visual draft to improve users' engagement and help scriptwriters rapidly iterate ideas.



To our best knowledge, we are the first to tackle the controllable script generation task, which controls a script's genre and storyline.
We also introduce a practical approach for plot-guided dialogue generation by treating the task as inverse dialogue summarization, which improves the diversity of the generated dialogue while maintaining relevancy to the plot.
In addition, we explore effective methods to produce an eloquent real-time visual presentation from the generated script.
According to the evaluation results, our scripts are controllable and preferred by humans.
Thus, \textbf{VScript} can serve as a tool for users to produce scripts with their preferences and for scriptwriters to optimize the writing process. We think \textbf{VScript} has the potential to promote AI-human collaboration in script generation.
Limitations and Ethical Considerations are discussed in Appendix \ref{appendix sec:Limitations} and \ref{appendix sec:Ethical}.


\section{Related Work}

\paragraph{Story Generation}
In recent years, methods for story generation have focused on using neural networks and have shown promising results.
\citet{martin2018event} decompose a story as a sequence of events and apply sequence modelling to generate the story.
\citet{fan-etal-2018-hierarchical} employ a hierarchical story generation by generating a premise and transforming it into a passage. Plan-and-Write~\citep{yao2019plan} extracts a storyline composed of keywords and generates a story based on the storyline. ~\citet{rashkin2019towards} generate a narrative using a set of phrases that describe key characters and events in a story. \citet{lovenia2022every} generate a story using a genre and an image as its context.

\paragraph{Controllable Text Generation Model}
Conditional deep generative models 
are effective in improving the controllability of the models. CTRL~\citep{keskar2019ctrl} is a class-conditional language model (CC-LM) pre-trained on 50 domains with different control codes. Plug and Play Language Models (PPLM)~\citet{dathathri2019plug} combine pre-trained language models and 
attribute classifiers to steer generation. GeDi~\citep{krause2020gedi} incorporates CC-LM as a discriminator to control generation towards the desired attribute. 


\section{Methodology}
As shown in Figure \ref{fig:system}, our framework can be decomposed into two modules: script generation and visual presentation. We will describe these two modules in detail in the following sections.


\subsection{Script Generation}
A plot, or so-called outline, is a vital component required for professional script writing, which specifies a series of headings showing the main themes that need to be discussed~\citep{owens2012video}. Following this concept, we first generate a genre-specific plot to hierarchically guide the dialogue and scene description generation. We condition the dialogue on the plot and the scene description on dialogue instead of the other way around since dialogues between characters change dynamically to reflect the progression of the plot, while scene descriptions mainly provide detailed information about where and how the dialogue takes place. In this work, we select four classic and popular genres for the plot generation, i.e., \texttt{Crime}, \texttt{Sci-Fi}, \texttt{War}, and \texttt{Romance}.

\begin{figure*}[!ht]
 \centering
 \includegraphics[width=0.98\linewidth]{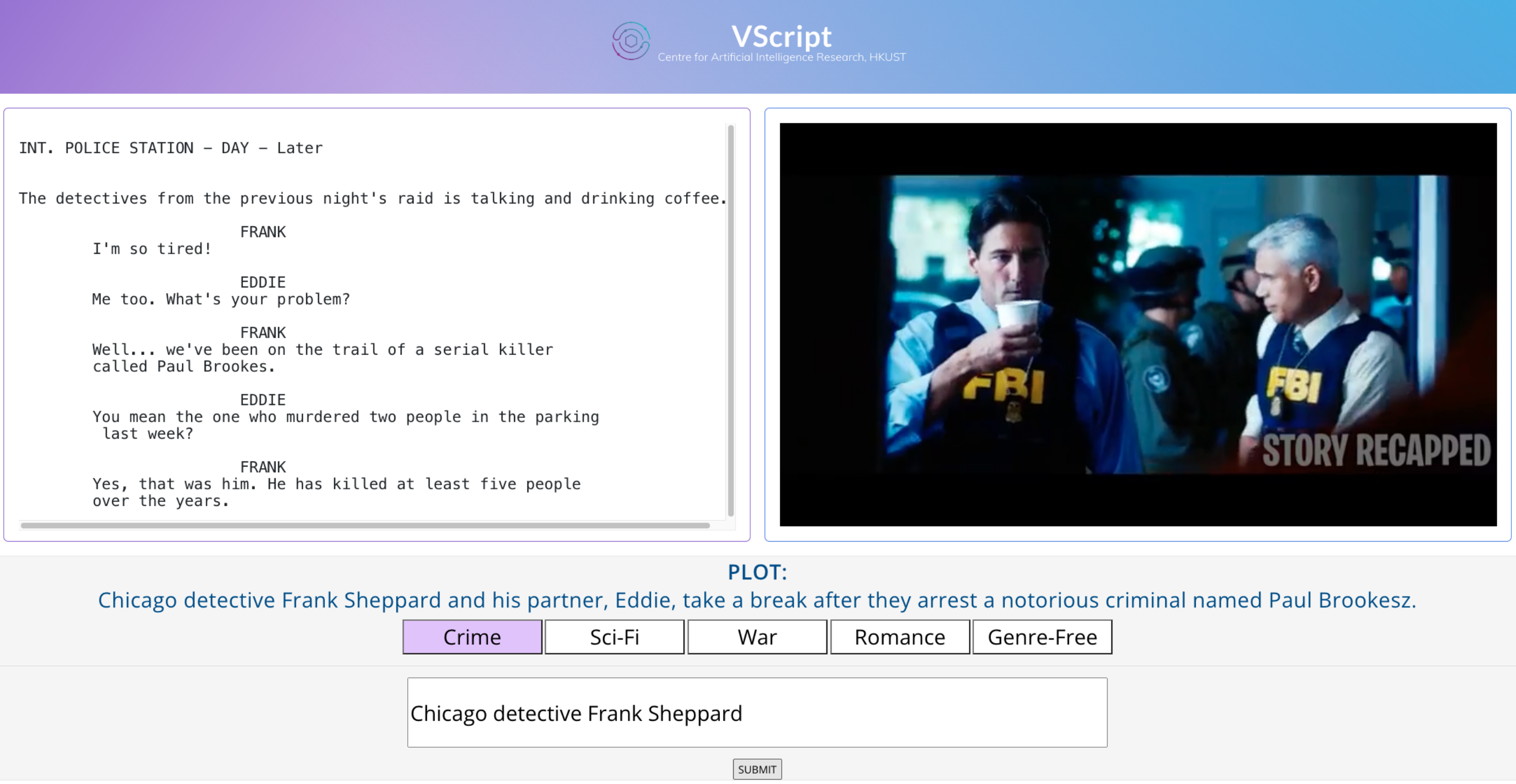}
  \caption{The \textbf{VScript}'s user interface. There are three main areas: script area (top left), video area (top right), and interaction area (bottom centre). For example, a user chooses ``\texttt{Crime}'', and input ``\texttt{Chicago detective Frank Sheppard}''. Then, the generated script and its visual presentation are displayed separately.}
  \label{fig:interface}
    \vspace{-1em}
\end{figure*}

\subsubsection{Genre-Specific Plot Generation}
\label{subsec:plot}
Inspired by CTRL~\citep{keskar2019ctrl}, we first train a class-conditional language model (CC-LM) for plot generation by adopting control codes which are a set of predefined genres. Then, by training different types of text with different control codes concatenated in the front, the model can learn the correlation between the types and the control codes so that the different control codes can guide the generation process of the language model. 
We fine-tune GPT2-large~\citep{radford2019language} on CMU Movie Summary Corpus\footnote{\url{http://www.cs.cmu.edu/~ark/personas/}\label{movie_genre}}, which contains 42,306 movie plots from Wikipedia and the corresponding metadata, such as genre. In our work, the genres of movies are treated as control codes. Each control code guides the generation of episodes of the desired type. See Appendix \ref{appendix sec:plot} for more details.



\paragraph{Plot Rescoring} To ensure the consistency of the generated plots and the target genre, we further train a multi-class genre classifier that can predict the probabilities of a plot belonging to each genre. Then, we generate $N$ plots with the same genre and starting words via Top-K sampling. Finally, we select the plot with the highest probability among all the generated $N$ plots.

\subsubsection{Plot-Guided Dialogue Generation}

The dialogue in the script is required to be casual, natural, and in line with the plot. However, to our knowledge, there is no open-source dataset for plot-to-dialogue generation, and building it would require intensive human labour. Thus, we treat this task as an inversed abstractive dialogue summarization, where the model is trained to generate the whole dialogue based on the dialogue summary. The model learns to generate the entire dialogue in one fell swoop, which is different from conventional dialogue models generating dialogues turn-by-turn. Two dialogue summarization corpora, SAMSum~\citep{gliwa2019samsum} and DialogSum Corpus~\citep{chen-etal-2021-dialogsum}, are combined as the training set. A GPT2-large model is trained on the inversed version of it.
During the inference time, we assume that each sentence in the plot can be expanded into a single scene, which can be decomposed into the scene description and the dialogue. We leverage our fine-tuned model to generate dialogues for each plot sentence.

\subsubsection{Scene Description Generation}
A scene description includes the scene header, i.e., location and time, and the scene context.
In order to infer such scene descriptions from each dialogue, we fine-tune the GPT2-large on a paired scene-dialogue corpus. 
During preprocessing on Film Corpus 2.0\footnote{\url{https://nlds.soe.ucsc.edu/fc2}\label{filmcorpus2}}, we pair each scene description with its corresponding dialogue to construct the dataset for dialogue-to-scene generation.
Finally, we concatenate the scene description and the corresponding dialogues to form a scene. A script is formed by concatenating multiple scenes. 

\subsection{Visual Presentation}
In addition to the generated script, we provide a visual presentation, which can serve as a rough visual draft for the users.
We retrieve a video clip whose caption describes similar actions or events to the generated script.
Note that we only utilize the visual contents of the retrieved video and ignore the auditory information. This disregard is intended to enhance retrieval quality, as the conversational content and visual appearance of a video are often inconsistent.
For example, a video shows two women sitting face to face at a café, looking bright and peaceful, while their conversation is about fierce interstellar wars.
In this section, we explain our video database construction process and the retrieval method.

\paragraph{Video Database Construction}
We construct a video database including news broadcasts, documentaries, and movie recaps from social media and only preserve the video if 1) it has captions; 2) there is a voice-over to introduce and describe what is happening; 3) most of the frames have rich content.
Post-processing and filtering are further conducted to ensure the quality of the videos.
The characters (including number and gender), time (day/night) and locations in the videos are detected for video retrieval. In addition, we classify the video captions by a zero-shot text classifier~\footnote{\url{https://huggingface.co/facebook/bart-large-mnli}} to split this database based on genres.
For more details, see Appendix~\ref{sec:videl-data-cons}. 
Since our method is zero-shot and independent of the video contents, users can replace the video database with any preferred videos.

\paragraph{Video Retrieval}
To match the video clips with the generated scripts, we use plots as queries for video retrieval since there may be trivial and lengthy dialogues in scripts that affect the coherence among the retrieved video clips.
For each plot sentence, we retrieve video clips from the video database by calculating the cosine similarity between the sentence embedding of the plot and the corresponding caption with the pre-trained DistilRoBERTa-based model~\footnote{\url{https://github.com/UKPLab/sentence-transformers}}.
We also use the videos' pre-detected gender and location information to filter out some improper candidates and select the best matching video clip.

\section{Interactive User Interface}
An example of the interaction between users and \textbf{VScript} is illustrated in Figure~\ref{fig:interface}.
The user interface comprises three parts: the script area (top left), the video area (top right), and the interaction area (bottom centre). First, users select the script genre among \texttt{Crime}, \texttt{Sci-Fi}, \texttt{War}, \texttt{Romance}, and \texttt{Genre-Free}. 
Second, users type the starting words into the input box and submit. Finally, the generated script will be displayed in the script area and its visual presentation in the video area. Users can interrupt at any time and choose the genre or input some words to steer the script's development.

\section{Experiments}

\subsection{Baselines}
\label{sec:baseline}
\paragraph{Plot Generation}
We fine-tune \textbf{GPT2}-large on CMU Movie Summary Corpus and a \textbf{CC-LM} using GPT2-large backbone on the same corpus without the genre-classifier.

\paragraph{Plot-Guided Dialogue Generation}
We fine-tune \textbf{DialoGPT}-large~\citep{zhang2020dialogpt} on the inversed SAMSum and DialogSum Corpus, where the model generates dialogue turn-by-turn iteratively.
We fine-tune GPT2-large on the inversed SAMSum and DialogSum Corpus, where the model generates dialogue turn-by-turn (\textbf{GPT2 T}).

\paragraph{Overall Script Generation}
In contrast to our proposed pipeline, we fine-tune GPT2-large directly on the scripts from the Film Corpus 1.0\footnote{\url{https://nlds.soe.ucsc.edu/fc}} in an end-to-end manner without plot \textbf{(GPT2\_E)}.

\paragraph{Video Retrieval}
To verify our video retrieval, we use \textbf{VideoCLIP}~\citep{xu2021videoclip}, a pre-trained model for zero-shot video and text understanding. 

\begin{table}[!t]
\centering
\setlength{\abovecaptionskip}{0.5em} 
\setlength{\belowcaptionskip}{-0.3em}
\resizebox{0.9\linewidth}{!}{ 
\begin{tabular}{@{}lcccccc@{}}
\toprule
\textbf{Model} & \textbf{PPL} & \textbf{Genre-ACC(\%)} \\ \hline %
GPT2 & 20.43 & - \\
CC-LM & 21.98 & 63.50 \\
CC-LM+Classifier (Ours) & 21.98 & 95.50 \\
\bottomrule
\end{tabular}
}
\caption{Automatic Evaluation for Plot Generation.}
\label{tab:plot}
\end{table}

\begin{table}[!t]
\centering
\setlength{\abovecaptionskip}{0.5em} 
\setlength{\belowcaptionskip}{-0.3em}
\resizebox{\linewidth}{!}{
\begin{tabular}{@{}lccccc@{}}
\toprule
\multicolumn{1}{c}{\textbf{Model}} & \textbf{BLEU} & \textbf{\begin{tabular}[c]{@{}c@{}}Sent\\Sim\end{tabular}} & \textbf{\begin{tabular}[c]{@{}c@{}}Repeat\\(\%)\end{tabular}} & \textbf{\begin{tabular}[c]{@{}c@{}}Dist-n \\ (n=1,2,3)\end{tabular}} \\ \hline %
DialoGPT & 13.35 & 54.18  & 20.25 & 1.7/13.6/42.69  \\
GPT2 T & 13.37 & 55.34 & 18.73 & 1.73/14.29/43.89  \\
\textbf{GPT2 (Ours)} & 16.4 & 58.97 & 9.68 & 3.19/22.4/53.32 \\
\bottomrule
\end{tabular}
}
\caption{Automatic Evaluation for Plot-Guided Dialogue Generation.}
\label{tab:dialog_result_supervised}
\end{table}

\begin{table}[!t]
\centering
\setlength{\abovecaptionskip}{0.5em} 
\setlength{\belowcaptionskip}{-0.3em}
\resizebox{0.75\linewidth}{!}{ 
\begin{tabular}{@{}lccc@{}}
\toprule
\multicolumn{1}{c}{\textbf{Model}} & \textbf{\begin{tabular}[c]{@{}c@{}}Dist-n (n=1,2,3)\end{tabular}} & \textbf{\begin{tabular}[c]{@{}c@{}}Repeat (\%)\end{tabular}}
\\ \midrule
GPT2\_E & 8.42/36.19/68.46 & 12.52 \\
\textbf{Ours} & 5.47/35.81/73.3  & 4.35 \\ 
\bottomrule
\end{tabular}
}
\caption{Automatic Evaluation for Scripts.}
\label{tab:script_unsuper}
\end{table}

\subsection{Evaluation}
\subsubsection{Automatic Evaluation}
\paragraph{Genre-Specific Plot Generation} We score perplexity (\textbf{PPL}) of texts generated from our model and baseline (GPT2-large) by another model for fluency evaluation.
We use the GPT-Neo-1.3B model since it is large enough to represent the real sentence distribution.
We also calculate \textbf{Genre-ACC}, the accuracy of genre control, with the NLI-based zero-shot text classifier.
As shown in Table \ref{tab:plot}, our method can control genre more effectively with only a slight reduction in fluency.

\paragraph{Plot-Guided Dialogue Generation} 
We evaluate models on the test set of SAMSum and DialogSum.
We use \textbf{BLEU} to compare the generated dialogue with the gold-standard human reference.
We also calculate \textbf{Sentence Similarity}, which is defined as the cosine similarity between sentence embeddings~\footnote{\url{https://huggingface.co/sentence-transformers/paraphrase-distilroberta-base-v2}} of plot and dialogue.
In addition, we calculate \textbf{Distinct-n} to measure the diversity of generated texts, and \textbf{Repeat}, the average percentage of the unigrams that occur in the previous 8 tokens~\citep{welleck2019neural}, to measure the level of repetition. 
As shown in Table \ref{tab:dialog_result_supervised}, generating the entire dialogue directly rather than turn-by-turn makes the plot-guided dialogues more similar to the gold references and higher semantic similarity with the plot. 
Both the generated dialogues and the scripts from our model (in Table~\ref{tab:script_unsuper}) show higher diversity and lower repetition over the baselines.

\subsubsection{Human Evaluation}
\label{subsec: human eval}
\begin{table}[!t]
\centering
\setlength{\abovecaptionskip}{0.5em} 
\setlength{\belowcaptionskip}{-0.3em}
\resizebox{0.8\linewidth}{!}{ 
\begin{tabular}{@{}ccccc@{}}
\toprule
\multicolumn{1}{c}{\textbf{Ours vs GPT2-E}} & \textbf{Win(\%)} & \textbf{Loss(\%)}& \textbf{Tie(\%)}\\ \midrule
\textbf{Preference} & 54.00 & 27.33 & 18.67 \\ 
\textbf{Genre Control} & 95.33 & 4.00 & 0.67 \\ 
\bottomrule
\end{tabular}
}
\caption{Human Evaluation for Scripts.}
\label{tab:human_script}
\end{table}

\begin{table}[!t]
\centering
\setlength{\abovecaptionskip}{0.5em} 
\setlength{\belowcaptionskip}{-1em}
\resizebox{0.9\linewidth}{!}{ 
\begin{tabular}{@{}ccccc@{}}
\toprule
\multicolumn{1}{c}{\textbf{Ours vs VideoCLIP}} & \textbf{Win(\%)} & \textbf{Loss(\%)}& \textbf{Tie(\%)}\\ \midrule
\textbf{Relevance} & 27.33& 13.33 & 59.33 \\ 
\bottomrule
\end{tabular}
}
\caption{Human Evaluation for Video Retrieval.}
\label{tab:human_video}
\end{table}

We conduct human evaluations further to assess the quality of \textbf{VScript} using Amazon Mechanical Turk.
We randomly select 50 samples per model, and three annotators then evaluate each sample to rule out potential bias.
We conduct A/B testing against the baseline GPT2\_E to assess generated scripts on \textbf{Preference} and \textbf{Genre Control}. 
For \textbf{Preference}, we ask the annotators to choose which script is the better one from three aspects~\footnote{Please refer to Appendix~\ref{appendix: human eval} for the results of each aspect.}: 1) format, whether the text meets the standard of film scripts; 2) fluency, whether the writing is smooth and grammatically correct; and 3) consistency, whether the content is logically consistent. For \textbf{Genre Control}, we ask the annotators to choose which script better belongs to a given genre.
In both tests, the annotators are given four choices: \{neither, both, sample A, or sample B\}.
As shown in Table \ref{tab:human_script}~\footnote{The result is statistically significant with p < 0.05.\label{pvalue}}, human judges prefer the scripts generated by \textbf{VScript}, which is inline with the automatic evaluation.
For video retrieval, we conduct A/B testing against VideoCLIP to evaluate the \textbf{Relevance}. The \textbf{Relevance} between the script and video retrieved by \textbf{VScript} is slightly higher than the baseline, as in Table \ref{tab:human_video} \footref{pvalue}. 



\section{Conclusion}
We propose the first controllable script generation framework \textbf{VScript} that can generate scripts of specific genres and follow the plots. Our framework adopts a hierarchical structure, which generates the plot, then the script and its visual presentation.
We adopt inversed abstractive summarization for dialogue generation.
Based on our experiments, \textbf{VScript} outperforms the baselines, and its effectiveness in genre control is proven.

\bibliography{main}
\bibliographystyle{acl_natbib}
\clearpage

\appendix

\section{Limitations}
\label{appendix sec:Limitations}
The performance of video retrieval is limited to some extent by the quality of the sentence embedding, face detection, and place detection via off-the-shelf tools.
The effect of visual presentation depends heavily on the quality and quantity of the video database.
More exploration in video retrieval or video generation can also improve the matching quality between a script and its visual presentation.
In addition, we would explore a more fine-grained control, such as specific settings, character personalities, or event details for future work. 

\section{Ethical Considerations}
\label{appendix sec:Ethical}
\paragraph{Copyright}
We collect publicly available YouTube videos using the official YouTube API and follow the typical processing procedures~\citep{ignat-etal-2021-whyact}. We use the muted video footage and are neutral to the opinions expressed therein.
We will not release the database and are accountable for violating other parties' rights or terms of service.

\paragraph{Toxic Content}
\textbf{VScript} leverages large language models, which raise awareness of carrying biases and toxic content~\citep{ousidhoum-etal-2021-probing}. Therefore, we create lists of banned words that block them from the generated script to filter the possible curse words, racial slurs and sexually explicit contents. 
Since our videos are retrieved from publicly accessible media, which could include bloody and erotic content, we also filter these video clips based on the descriptions and captions.

\section{Genre-Specific Plot Generation}
\label{appendix sec:plot}

\begin{figure}[th]
 \centering
 \includegraphics[width=1\linewidth]{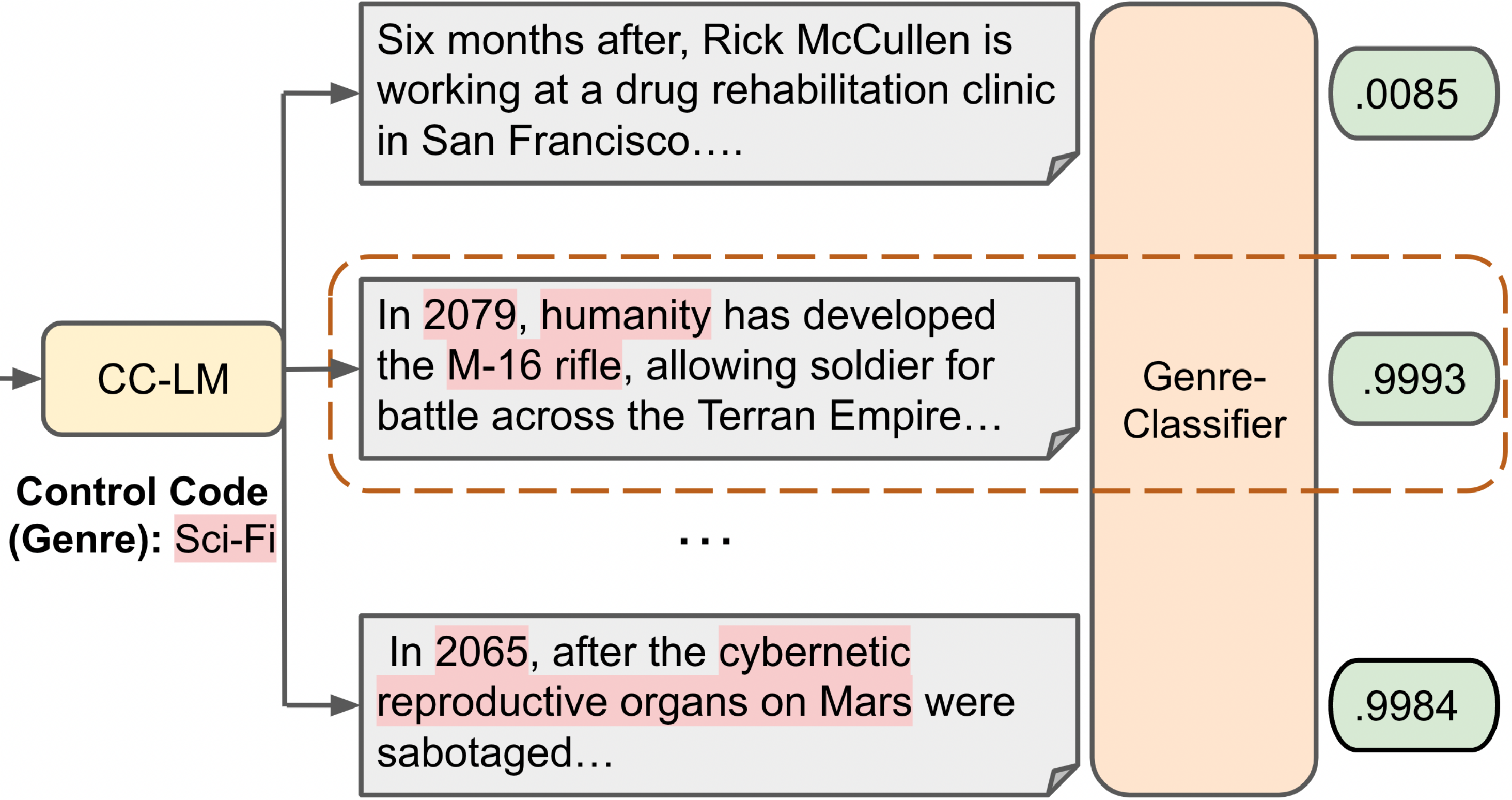}
  \caption{The process of plot generation. The CC-LM generates several candidates, and then the Genre-Classifier scores and ranks them to select the one that most belongs to the expected genre.}
  \label{fig:plot}
\end{figure}

As shown in Figure \ref{fig:plot}, the genres of movies are treated as control codes. Each control code guides the generation of episodes of the desired type.
The generation probability distribution can be decomposed as follows:
\begin{equation}
\setlength\abovedisplayskip{3pt}
\setlength\belowdisplayskip{3pt}
p(x|c^g)=\prod _{t=1}^T p(x_t|x_{<t},c^g)
\label{eq:CCLM_p}
\end{equation}
$\mathbf{C} = \left \{ c^1,...,c^g,...,c^G \right \}$ denotes the control code, where $c^g$ means the control code for g-th genre (G genres in total).

The CC-LM is trained on a set of plots $\left \{ x_{1:T^1}^1,...,x_{1:T^n}^n,...,x_{1:T^N}^N \right \}$, where each plot $x_{1:T^n}^n$ corresponds with the control code $c^x \in \mathbf{C}$. The training loss is denoted as:
\begin{equation}
\setlength\abovedisplayskip{3pt}
\setlength\belowdisplayskip{3pt}
L=-\sum _{n=1}^{N}\sum _{t=1}^{T^n}\log p_\theta (x_t^n|x^n_{<t},c^x)
\label{eq:CCLM_L}
\end{equation}

\paragraph{Plot Rescoring} To verify the genres of the generated plots, we train a multi-class genre classifier $\phi$ to predict the probability of the generated plot belonging to a specific genre $g$. By leveraging Top-K sampling, we generate $N (N=10)$ plots $\{ X_1,...,X_n,...X_N\}$ with the genre and the starting words. Finally, we select the plot $X_g^*$ with the highest probability over all the generated plots, which is defined as:
\begin{equation}
\setlength\abovedisplayskip{3pt}
\setlength\belowdisplayskip{3pt}
X_{g}^* = \mathop{\arg\max}\limits_{n\in \mathbf{N}} p_{\phi}(\mathbf{y}=y_g|X_n)
\end{equation}

where $\mathbf{y}=\{ y_1,...,y_g,...y_G\}$ denotes the genre classes of the generated plot and $y_g$ is the class corresponding to genre $g$.

\section{Video Database Construction}
\label{sec:videl-data-cons}
Firstly, we obtain videos and corresponding captions about news broadcasts and movie recaps from YouTube via the official API. 
Since some captions are generated automatically word by word, there is no punctuation, and they are not spliced into sentences. We use DeepSegment\footnote{\url{https://github.com/notAI-tech/deepsegment}} to sentence-tokenize these captions.
To ensure the richness of the video image content and improve the quality of video clips, the frames only consisting of hosts or speakers are filtered automatically. We use RetinaFace-R50 from InsightFace \footnote{\url{https://github.com/deepinsight/insightface}} to detect face. 
If there is only one face in the centre of the picture whose size is within an appropriate range, and it does not move for several frames, we will judge it as a speaker and delete these frames. 
We also use InsightFace to detect the gender of characters in the videos per second.
We use DenseNet-161 from Places365 \footnote{\url{https://github.com/CSAILVision/places365}} to recognize the location of scene in the video.
Furthermore, in order to match the time in scene description, we train a Vision Transformer (ViT)~\footnote{\url{https://huggingface.co/google/vit-base-patch16-224-in21k}}-based day/night image classifier on Aachen Day-Night Dataset~\footnote{\url{https://www.visuallocalization.net/}}, AMOS Day-Night Dataset~\footnote{\url{https://www.kaggle.com/datasets/stevemark/daynight-dataset}}, and ~\footnote{\url{https://github.com/kushagra2jindal/DayNightClassificationModel}}.
Finally, we classify the video captions by NLI-based Zero Shot Text Classifier~\citep{yin2019benchmarking} to split this corpus based on genres.
Since our method is zero-shot and independent of the video contents, users can update or replace the video database on their wish.

\section{Background Music}
\label{sec:music}
In order to render the atmosphere, we also use different styles and moods of music for different genres of the script.
For example, the music for crime is rapid and intense, while that for romance is relaxing and soothing.

\section{Experimental Settings}
\label{sec:exp-setting}

\subsection{Plot Generation}
\label{appendix:plot}
For CC-LM, we fine-tune GPT2-large with control codes (prefixes): ``\texttt{This is a crime/romance/sci-fi/war plot.}''.
We use the training hyperparameters: the learning rate is 3e-5, AdamW optimizer, and WarmupDecayLR scheduler and generate plots using top-k (k=4) sampling.
For Genre-Classifier, we fine-tune BART-large with the same training hyperparameters: the learning rate is 3e-5, AdamW optimizer, and WarmupDecayLR scheduler.
For the GPT2 baseline, we fine-tune the model with the same hyperparameter setting as GPT2-large models in our pipeline.

\begin{table}[!t]
\centering
\resizebox{\columnwidth}{!}{ 
\begin{tabular}{@{}lcccc@{}}
\toprule
\multicolumn{1}{c}{\textbf{Model}} & \textbf{Format (\%)} & \textbf{Fluency} & \textbf{Consistency} \\ \midrule
GPT2-E & 93(0.26) & 2.92(0.45) & 2.85(0.45)\\ 
\textbf{VScript} & \textbf{95(0.22)} & \textbf{3.06(0.42)} & \textbf{3.00(0.50)}\\
\bottomrule
\end{tabular}
}
\caption{Additional human evaluation for scripts.}
\label{tab:human2}
\end{table}

\subsection{Plot-Guided Dialogue Generation}
The model in this stage of our pipeline has the same training and generation hyperparameters as the GPT2-large model in Appendix~\ref{appendix:plot}.
For the DialogGPT baseline, we fine-tune the model with a learning rate (5e-5), and the other hyperparameters are the same as GPT2-large models in our pipeline.
For GPT2 Turn-by-Turn (GPT2 T) baseline, we use the same hyperparameter setting as the GPT2-large models in our pipeline.

\subsection{Scene Description Generation}
The model in this stage of our pipeline has the same hyperparameters as GPT2-large model in Appendix~\ref{appendix:plot}.

\subsection{Overall Script Generation}
For GPT2\_E baseline, we use the same hyperparameter set with the GPT2-large model in Appendix~\ref{appendix:plot}.

\section{Additional Human Evaluation}
\label{appendix: human eval}
In addition, we also evaluate the quality of each generated script for a more comprehensive study, compared with GPT2-E model. As mentioned in Section~\ref{subsec: human eval}, we breakdown the \textbf{Preference} metric into three aspects: \textbf{Format}, \textbf{Fluency}, and \textbf{Consistency}. \textbf{Format} measures whether our generation meets the standard of the script, which is defined as a text that contains a scene header, and dialogue (including monologue). For this metric, we conduct True/False binary evaluation. 
\textbf{Fluency} reflects whether the writing is smooth and non-repetitive, without grammatical and spelling mistakes. \textbf{Consistency} emphasizes whether the content is logically consistent. 
For Fluency and Consistency, we leverage a 4-point Likert Scale, where 1 indicates non-fluent/inconsistent and 4 indicates a very fluent/consistent text. 
For each model, we randomly sample 50 generated scripts from each model. And each script is evaluated by three annotators. 
An individual t-test is conducted for significance validation of the human evaluation results. 
As shown in Table~\ref{tab:human2}~\footnote{The result is statistically significant with p < 0.05.}, while achieving script formulation, our system has slightly higher fluency than the baseline model, indicating that fluency is not compromised.
Our framework is also more consistent and logical, with considerably higher consistency than GPT2-E. 

\end{document}